\pdfoutput=1

\documentclass[11pt]{article}

\usepackage{naacl2021}

\usepackage{times}
\usepackage{latexsym}

\usepackage{algorithm}
\usepackage{algorithmic}
\usepackage{multirow}
\usepackage{graphicx}
\usepackage{amsmath}
\usepackage{subcaption}
\usepackage{enumitem}
\usepackage{booktabs}

\usepackage[T1]{fontenc}

\usepackage[utf8]{inputenc}

\usepackage{microtype}

\title{RocketQA: An Optimized Training Approach to Dense Passage Retrieval for Open-Domain Question Answering
}

\author{ \textbf{Yingqi Qu\textsuperscript{1}, Yuchen Ding\textsuperscript{1}, Jing Liu\textsuperscript{1}\thanks{\llap{}\:\:\:Corresponding authors. }, Kai Liu\textsuperscript{1}, Ruiyang Ren\textsuperscript{2}}\thanks{\textsuperscript{\dag} The work was done when Ruiyang Ren was doing internship at Baidu. } \\
 \textbf{Wayne Xin Zhao\textsuperscript{2}\footnotemark[1], Daxiang Dong\textsuperscript{1}, Hua Wu\textsuperscript{1} and Haifeng Wang\textsuperscript{1}} \\
	\textsuperscript{1}Baidu Inc.; \textsuperscript{2}Gaoling School of Artificial Intelligence, Renmin University of China\\
	\{quyingqi, dingyuchen, liujing46, liukai20, dongdaxiang, 
	wu\_hua, wanghaifeng\}@baidu.com \\
	reyon.ren@ruc.edu.cn, batmanfly@gmail.com}

\begin{document}
\maketitle
\begin{abstract}
  In open-domain question answering, dense passage retrieval has become a new paradigm to retrieve relevant passages for finding answers. 
  Typically,  the dual-encoder architecture is  adopted to learn  dense representations of questions and passages for semantic matching.
  However, it is difficult to effectively train a dual-encoder due to the challenges including the discrepancy between training and inference, the existence of unlabeled positives and limited training data.
  To address these challenges, we propose an optimized training approach, called \emph{RocketQA}, to improving dense passage retrieval. We make three major technical
   contributions in RocketQA, namely cross-batch negatives, denoised hard negatives and data augmentation. 
   The experiment results show that RocketQA significantly outperforms previous state-of-the-art models on both MSMARCO and Natural Questions. 
   We also conduct extensive experiments to examine the effectiveness of the three strategies in RocketQA. Besides, we demonstrate that the performance of end-to-end QA can be improved based on our RocketQA retriever~\footnote{Our code is available at ~\url{https://github.com/PaddlePaddle/Research/tree/master/NLP/NAACL2021-RocketQA}}. 
   
   \end{abstract}

\section{Introduction} \label{introduction}
Open-domain question answering (QA) aims to find the answers to natural language questions from a large collection of documents. 
Early QA systems \citep{brill2002analysis,dang2007overview,ferrucci2010building} constructed complicated pipelines consisting of multiple components, including question understanding, document retrieval, passage ranking and answer extraction. Recently, inspired by the advancements of machine reading comprehension (MRC), \citet{chen2017reading} proposed a simplified two-stage approach, where a traditional IR \textit{retriever}~(e.g., TF-IDF or BM25)  first selects a few relevant passages as contexts, and then a neural \textit{reader} reads the contexts and extracts the answers. As the recall component, the first-stage retriever  significantly affects the final QA performance. Though efficient with an inverted index, traditional IR retrievers with term-based sparse representations have limited capabilities  in matching questions and passages, e.g., term mismatch. 

\begin{figure}[!tb]
\centering

\begin{subfigure}{0.45\textwidth}
\centering
\includegraphics[width=0.90\linewidth]{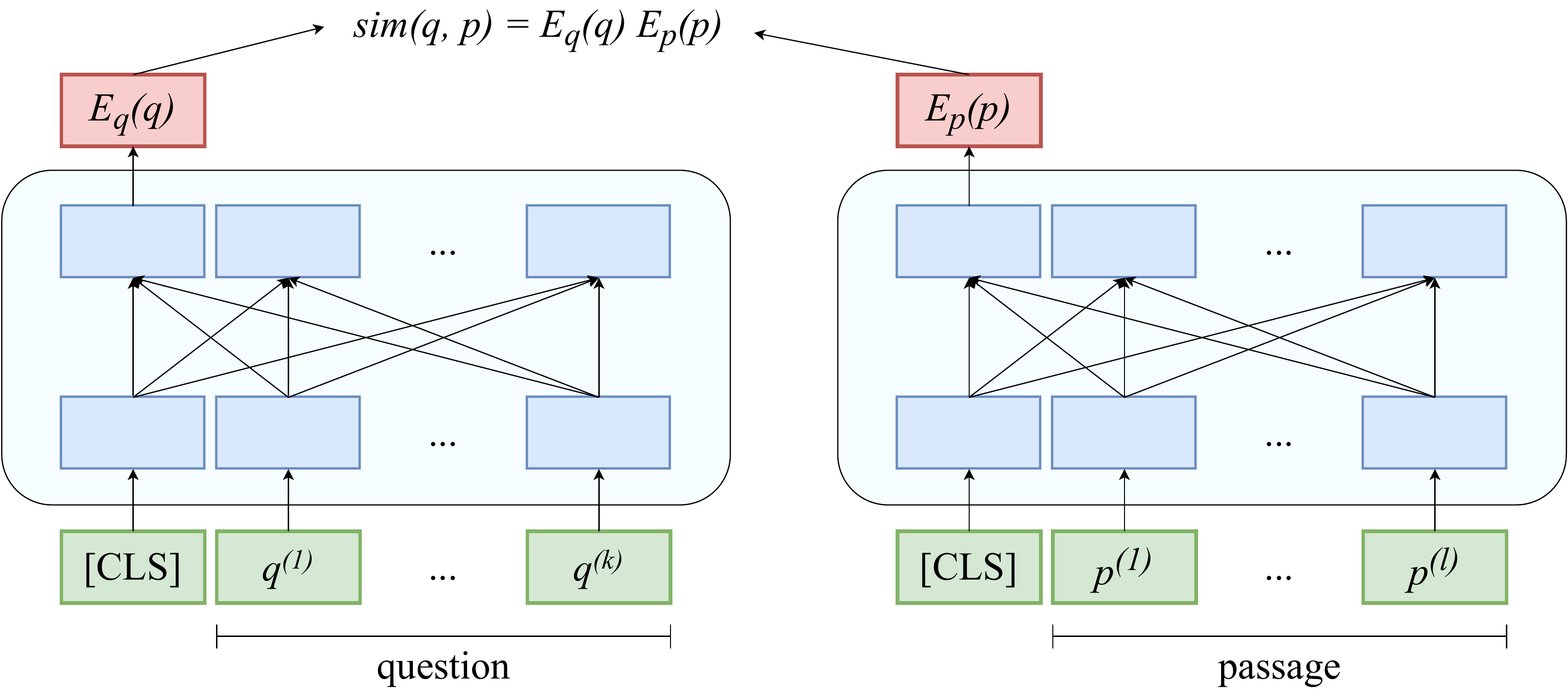}
\caption{A dual-encoder based on pre-trained LMs.}
\label{fig-dual-encoder}
\end{subfigure}

\begin{subfigure}{0.45\textwidth}
\centering
\includegraphics[width=0.90\linewidth]{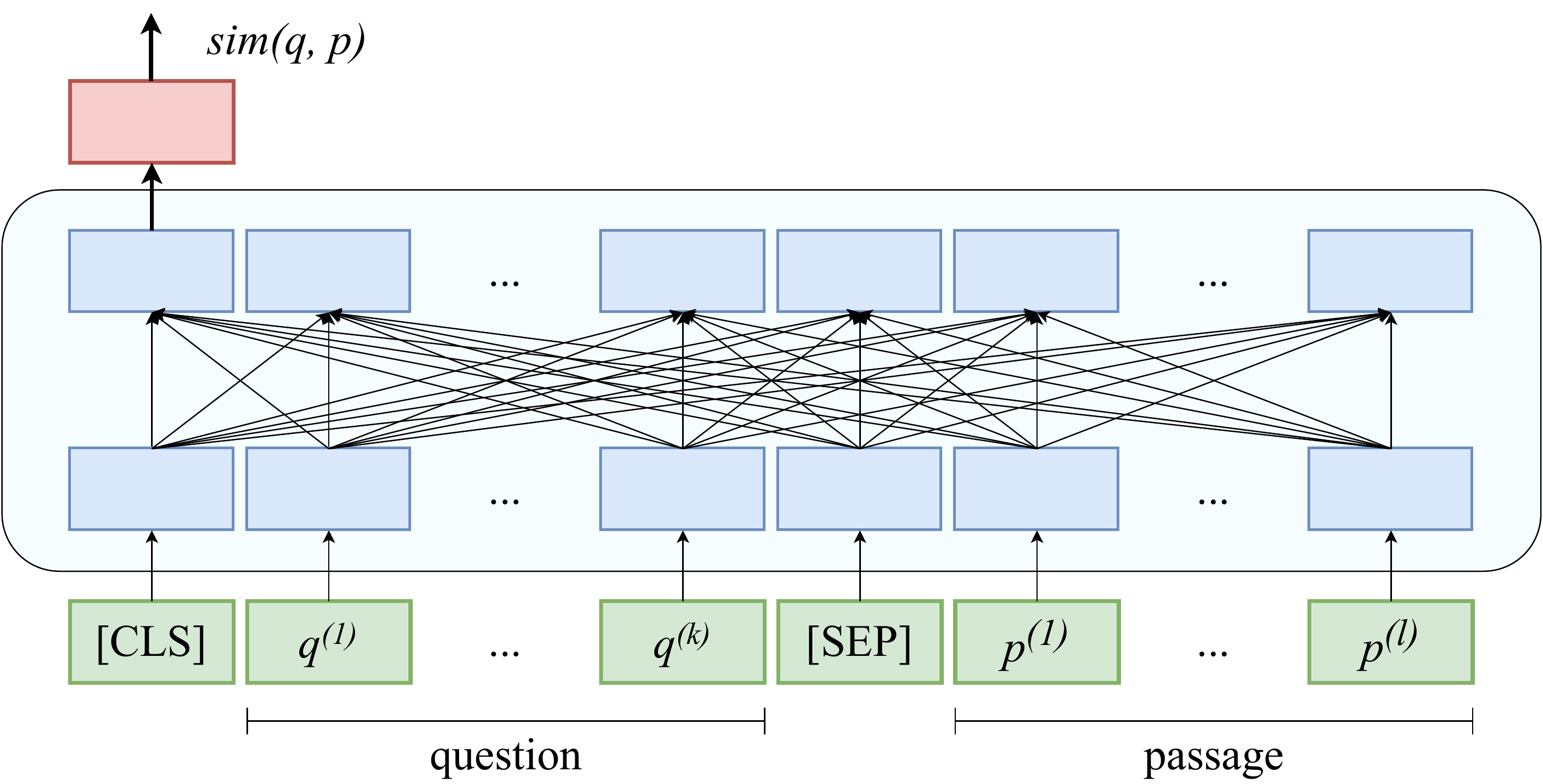}
\caption{A cross-encoder based on pre-trained LMs.}
\label{fig-cross-encoder}
\end{subfigure}
\vspace{-2mm}
\caption{The comparison of dual-encoder and cross-encoder architectures.}
\label{fig-encoder-architecture}
\vspace{-6mm}
\end{figure}

To deal with the issue of term mismatch, the dual-encoder architecture (as shown in Figure \ref{fig-dual-encoder}) has been widely explored \citep{lee2019latent,guu2020realm,karpukhin2020dense,luan2020sparse,xiong2020approximate} to learn dense representations of questions and passages in an end-to-end manner, which provides better representations for semantic matching. 
These studies first separately encode questions and passages to obtain their dense representations, and then compute the similarity between the dense representations using similarity functions such as cosine or dot product. 
Typically, the dual-encoder is trained by using in-batch random negatives: for each \emph{question}-\emph{positive passage} pair in a training batch, the positive passages for the other questions in the batch would be used as negatives. 
However, it is still difficult to effectively train a dual-encoder for dense passage retrieval due to the following three major challenges. 

First, there exists the discrepancy between training and inference for the dual-encoder retriever.
During inference, the retriever needs to identify positive (or relevant) passages for each question from a large collection containing millions of candidates. However, during training, the model 
is learned to estimate the probabilities of positive passages in a small candidate set for each question, due to the limited memory of a single GPU (or other device). To reduce such a discrepancy, previous work tried to design specific mechanisms for selecting a few hard negatives from the top-$k$ retrieved candidates~\citep{gillick2019learning,wu2019zero,karpukhin2020dense,luan2020sparse,xiong2020approximate}. However, it suffers from the false negative issue due to the following challenge. 

Second, there might be a large number of unlabeled positives. Usually, it is  infeasible to completely annotate all the candidate passages for one question. 
By only examining the the top-$K$ passages retrieved by a specific retrieval approach (e.g. BM25), the annotators are likely to miss relevant passages to a question. Taking the MSMARCO dataset~\citep{nguyen2016ms} as an example, each question has only $1.1$ annotated positive passages on average, while there are $8.8$M passages in the whole collection. As will be shown in our experiments, we manually examine the top-retrieved passages that were not labeled as positives in the original MSMARCO dataset, and we find that $70\%$ of them are actually positives. Hence, it is likely to bring false negatives when sampling hard negatives from the top-$k$ retrieved passages. 

Third, it is expensive to acquire large-scale training data for open-domain QA. MSMARCO and Natural Questions~\cite{kwiatkowski2019natural} are two largest datasets for open-domain QA. They are created from commercial search engines, and have 516K and 300K annotated questions, respectively. 
However, it is still insufficient to cover all the topics of questions issued by users to search engines. 

In this paper, we focus on addressing these challenges so as to effectively train a dual-encoder retriever for open-domain QA. We propose an optimized training approach, called \emph{RocketQA}, to improving dense passage retrieval. Considering the above challenges, we make three major technical contributions in RocketQA. First, RocketQA introduces cross-batch negatives. Comparing to in-batch negatives, it increases the number of available  negatives for each question during training, and alleviates the discrepancy between training and inference. 
Second, RocketQA introduces denoised hard negatives.
It aims to remove false negatives from the top-ranked results retrieved by a retriever, and derive more reliable hard negatives. 
Third, RocketQA leverages large-scale unsupervised data ``\emph{labeled}'' by a cross-encoder (as shown in Figure \ref{fig-cross-encoder}) for data augmentation. 
Though inefficient, the cross-encoder architecture has been found to be more capable than the dual-encoder architecture in both theory and practice~\citep{luan2020sparse}. 
Therefore, we utilize a cross-encoder to generate high-quality pseudo labels for unlabeled data which are used to train the dual-encoder retriever. 
The contributions of this paper are as follows: 
\begin{itemize}[topsep=0pt,parsep=0pt,partopsep=0pt]
	\item The proposed RocketQA introduces three novel training strategies
	to improve dense passage retrieval for open-domain QA, namely cross-batch negatives, 
	denoised hard negatives, and data augmentation.
	\item The overall experiments show that our proposed RocketQA significantly outperforms previous state-of-the-art models on both MSMARCO and Natural Questions datasets. 
	\item We conduct extensive experiments to examine the effectiveness of the above three strategies in RocketQA. Experimental results show that the three strategies are effective to improve the performance of dense passage retrieval. 
	\item We also demonstrate that the performance of end-to-end QA can be improved based on our RocketQA retriever. 
\end{itemize}

 \vspace{-2mm}
\section{Related Work} \vspace{-2mm}

\textbf{Passage retrieval for open-domain QA} 
For open-domain QA, a passage retriever is an important component to identify relevant passages for answer extraction. Traditional approaches~\citep{chen2017reading} implemented term-based passage retrievers (e.g. TF-IDF and BM25), which have limited representation capabilities.
 Recently, researchers have utilized deep learning to improve traditional passage retrievers, including document expansions~\citep{nogueira2019document}, question expansions~\citep{mao2020generation} and term weight estimation~\citep{dai2019deeper}. 
 
Different from the above term-based approaches, dense passage retrieval has been proposed to represent both questions and documents as dense vectors (i.e., embeddings), typically in a dual-encoder architecture (as shown in Figure \ref{fig-dual-encoder}). Existing approaches can be divided into two categories: (1) self-supervised pre-training for retrieval~\citep{lee2019latent,guu2020realm,chang2020pre} and (2) fine-tuning pre-trained language models on labeled data.  Our work follows the second class of approaches, which show better performance with less cost. Although the dual-encoder architecture enables the appealing paradigm of dense retrieval, it is difficult to effectively train a retriever with such an architecture. As discussed in Section ~\ref{introduction}, it suffers from a number of challenges, including the  training and inference discrepancy, a large number of unlabeled positives and limited training data. Several recent studies \citep{karpukhin2020dense,luan2020sparse,chang2020pre,Henderson2017EfficientNL}  tried to address the first challenge by designing complicated sampling mechanism to generate hard negatives. However, it still suffers from the issue of false negatives. The later two challenges have seldom been considered for open-domain QA. 

\textbf{Passage re-ranking for open-domain QA} 
Based on the retrieved passages from a first-stage retriever, 
BERT-based rerankers have recently been applied to retrieval-based question answering and search-related tasks \citep{wang2019multi,nogueira2019passage,nogueira2019multi,yan2019idst}, and yield substantial improvements over the traditional methods. 
Although effective to some extent, these rankers employ the cross-encoder architecture (as shown in Figure \ref{fig-cross-encoder}) that is impractical to be applied to all passages in a corpus with respect to a question. The re-rankers~\citep{khattab2020colbert,Gao2020ModularizedTR} with light weight interaction based on the representations of dense retrievers have been studied. However, these techniques still rely on a separate retriever which provides candidates and representations. As a comparison, we focus on developing dual-encoder based retrievers.

\vspace{-2mm}
\section{Approach} \vspace{-1mm}
In this section, we propose an optimized training approach to dense passage retrieval for open-domain QA, namely \emph{RocketQA}. We first introduce the background of the dual-encoder architecture, and then describe the three novel training strategies in RocketQA. Lastly, we present the whole training procedure of RocketQA. 

\subsection{Task Description}
The task of open-domain QA is described as follows. Given a natural language question, a system is required to answer it based on a large collection of documents. Let $C$ denote the corpus, consisting of $N$ documents. We split the $N$ documents into $M$ passages, denoted by $p_1$, $p_2$, ..., $p_M$, where each passage $p_i$ can be viewed as an $l$-length sequence of tokens $p_i^{(1)}$, $p_i^{(2)}$, ..., $p_i^{(l)}$.
Given a question $q$, the task is to find a passage $p_i$ among the $M$ candidates, and extract a span $p_i^{(s)}$, $p_i^{(s+1)}$, ..., $p_i^{(e)}$ from $p_i$ that can answer the question. In this paper, we mainly focus on developing a dense retriever to retrieve the passages that contain the answer. 

\subsection{The Dual-Encoder Architecture}
We develop our passage retriever based on the typical dual-encoder architecture, as illustrated in Figure~\ref{fig-dual-encoder}.  
First, a dense passage retriever uses an encoder $E_p (\cdot)$ to obtain the $d$-dimensional real-valued vectors (a.k.a., embedding) of passages. Then, an index of passage embeddings is built for retrieval. At query time, another encoder $E_q (\cdot)$ is applied to embed the input question to a $d$-dimensional real-valued vector, and $k$ passages whose embeddings are the closest to the question's will be retrieved. The similarity between the question $q$ and a candidate passage $p$ can be computed as the dot product of their vectors:
\vspace{-2mm}
\begin{equation}
	\text{sim}(q, p) = E_q (q) \cdot E_p (p).
\vspace{-1mm}
\end{equation}

In practice, the separation of question encoding and passage encoding is desirable, so that the dense representations of all passages can be pre-computed for efficient retrieval. Here, we adopt two independent neural networks initialized from pre-trained LMs for the two encoders $E_q (\cdot)$ and $E_p (\cdot)$ separately, and take the representations at the first token (e.g., \textsc{[CLS]} symbol in BERT) as the output for encoding.

\textbf{Training} The training objective is to learn dense representations of questions and passages so that \emph{question}-\emph{positive passage} pairs have higher similarity than the \emph{question}-\emph{negative passage} pairs in training data. Formally, given 
a question $q_i$ together with its positive passage $p_i^+$  and $m$  negative passages $\{p_{i,j}^-\}_{j=1}^m$,
we minimize the loss function:

\begin{equation}\label{eq-L}
\begin{split}
			&\mathcal{L}(q_i, p_i^+, \{p_{i,j}^-\}_{j=1}^m)  \\
			=&-\log \frac{e^{\text{sim}(q_i, p_i^+)}}{e^{\text{sim}(q_i, p_i^+)} + \sum_{j=1}^{m} e^{\text{sim}(q_i, p_{i,j}^-)}},
\end{split}
\end{equation}
where we aim to optimize the negative log likelihood of the positive passage against a set of $m$ negative passages. Ideally, we should take all the negative passages in the whole collection into consideration in Equation~\ref{eq-L}.
However, it is computationally infeasible to consider a large number of negative samples for a question, and hence $m$ is practically set to a small number that is far less than $M$. As what will be discussed later, both the number and the quality of negatives affect the final performance of passage retrieval. 

\textbf{Inference} In our implementation, we use FAISS~\citep{johnson2019billion} to index the dense representations of all passages. Specifically, we use IndexFlatIP for indexing and the exact maximum inner product search for querying. 

\subsection{Optimized Training Approach}
In Section ~\ref{introduction}, 
we have discussed three major challenges in training the dual-encoder based retriever, including the training and inference discrepancy, the existence of unlabeled positives, and limited training data.
Next, we propose three improved training strategies to address the three challenges.  

\begin{figure}[t]
	\centering 
	\includegraphics[width=0.42\textwidth]{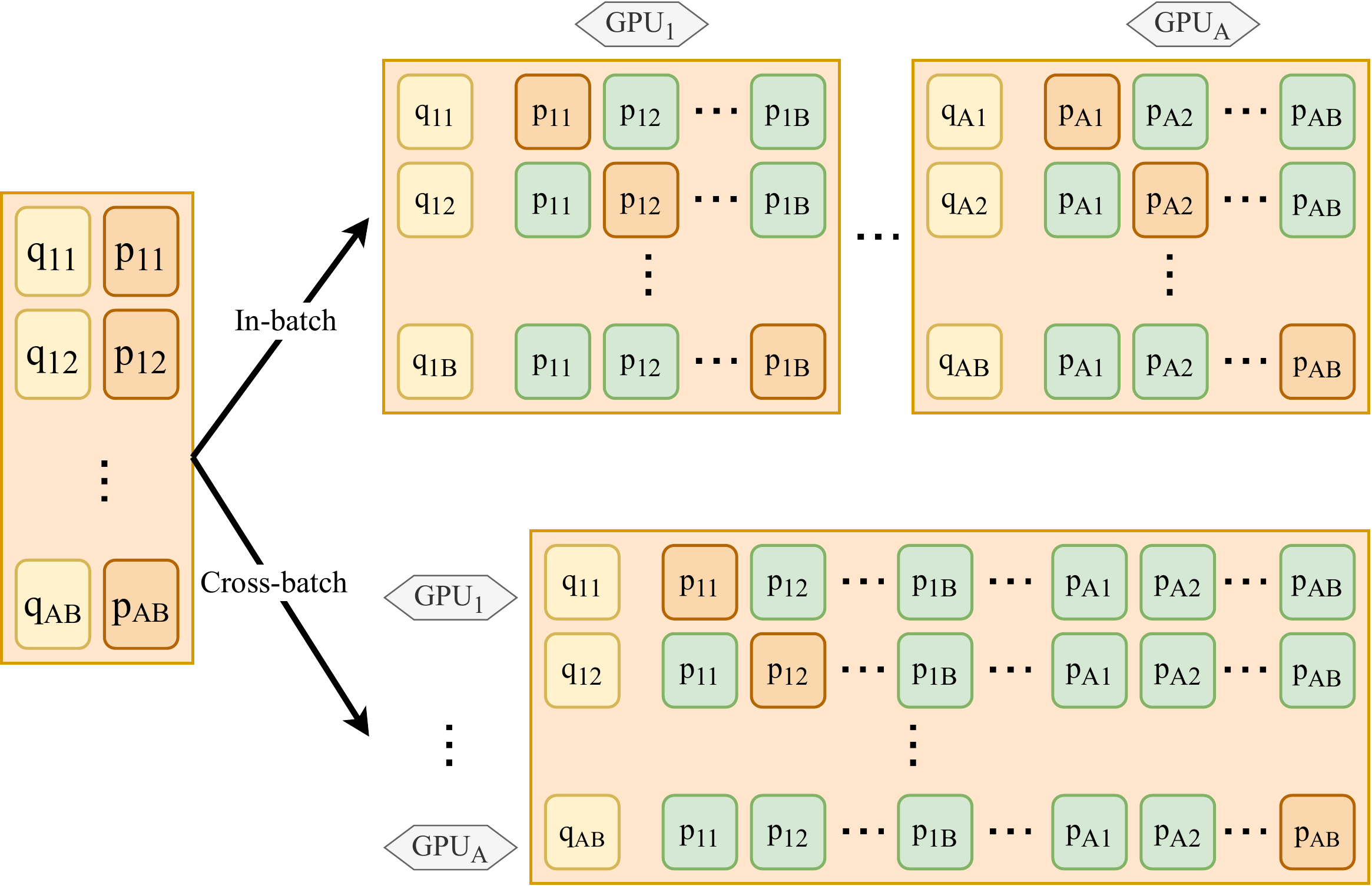}
	\caption{The comparison of traditional in-batch negatives and our cross-batch negatives when trained on multiple GPUs, where $A$ is the number of GPUs, and $B$ is the number of questions in each min-batch. } 
	\label{fig-cross-batch}
\vspace{-3mm}
\end{figure}

\textbf{Cross-batch Negatives} When training the dual-encoder, the trick of in-batch negatives has been widely used in previous work \citep{Henderson2017EfficientNL,gillick2019learning,wu2019zero,karpukhin2020dense,luan2020sparse}. 
Assume that there are $B$ questions in a mini-batch on a single GPU, and each question has one positive passage. With the in-batch negative trick, each question can be further paired with $B - 1$ negatives (i.e., positive passages of the rest questions) without sampling additional negatives. In-batch negative training is a memory-efficient way to reuse the examples already loaded in a mini-batch rather than sampling new negatives, which increases the number of negatives for each question. 
As illustrated at the top of Figure~\ref{fig-cross-batch}, we present an example for in-batch negatives when training on $A$ GPUs in a data parallel way. 
To further optimize the training with more negatives, we propose to use cross-batch negatives when training on multiple GPUs, as illustrated at the bottom of Figure~\ref{fig-cross-batch}. 
Specifically, we first compute the passage embeddings within each single GPU, and then share these passage embeddings among all the GPUs. Besides the in-batch negatives, we collect all passages (i.e., their dense representations) from other GPUs as the additional negatives for each question. Hence, with $A$ GPUs (or mini-batches)~\footnote{Note that cross-batch negatives can be applied in both settings of single-GPU and multi-GPUs. When there is only a single GPU available, it can be implemented in an accumulation way while trading off training time.}, we can indeed obtain $A \times B - 1$ negatives for a given question, which is approximately $A$ times as many as the original number of in-batch negatives. 
In this way, we can use more negatives in the training objective of Equation~\ref{eq-L}, so that the results are expected to be improved. 

\begin{figure*}[ht]
	\centering 
	\includegraphics[width=0.9\textwidth]{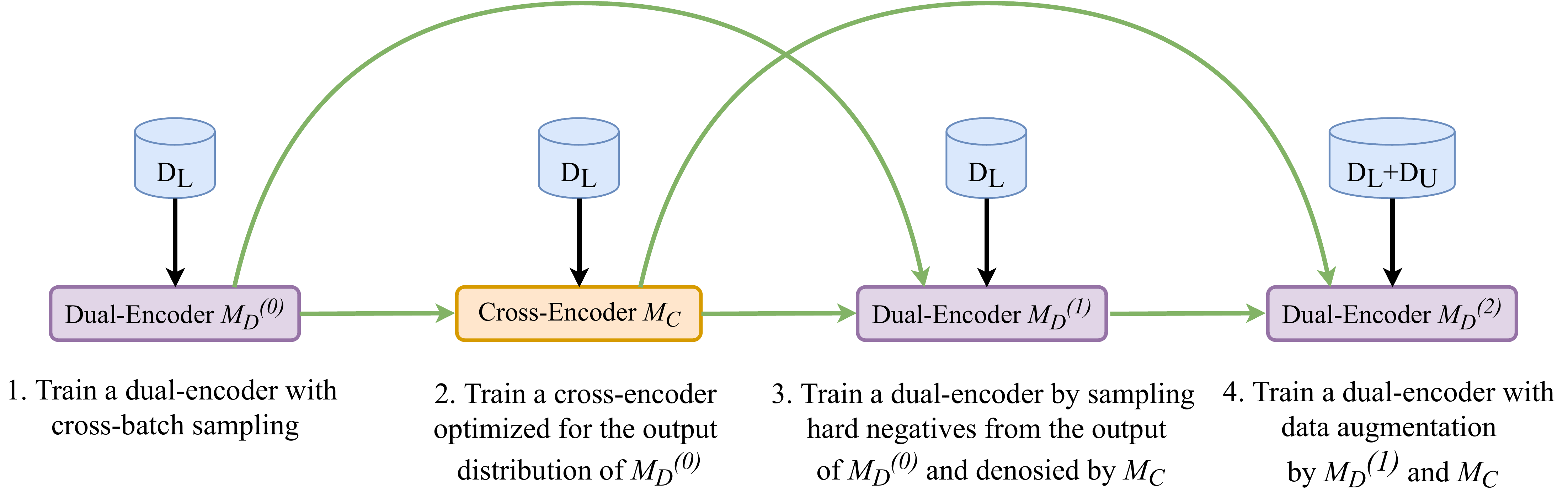}
	\caption{The pipeline of the optimized training approach RocketQA. $M_D$ and $M_C$ denote the dual-encoder and cross-encoder, respectively. We use $M_D^{(0)}$, $M_D^{(1)}$  and $M_D^{(2)}$ to denote the learned dual-encoders  after different steps.  }
	\label{fig-model} 
\vspace{-3mm}
\end{figure*}

\textbf{Denoised Hard Negatives} Although the above strategy can increase the number of negatives, most of negatives are easy ones, which can be easily discriminated. While, hard negatives are shown to be important to train a dual-encoder~\citep{gillick2019learning,wu2019zero,karpukhin2020dense,luan2020sparse,xiong2020approximate}. 
To obtain hard negatives, a straightforward method is to select the top-ranked passages (excluding the labeled positive passages) as negative samples. 
However, it is likely to bring false negatives (i.e., unlabeled positives), since the annotators can only annotate a few top-retrieved passages (as discussed in Section~\ref{introduction}).
Another note is that previous work mainly focuses on factoid questions, to which the answers are short and concise. Hence, it is not challenging to filter false negatives by using the short answers~\citep{karpukhin2020dense}. However, it cannot apply to non-factoid questions. In this paper, we aim to learn dense passage retrieval for both factoid questions and non-factoid questions, which needs a more effective way for denoising hard negatives. 

Here, our idea is to utilize a well-trained cross-encoder to remove top-retrieved passages that are likely to be false negatives. Because the cross-encoder architecture is more powerful for capturing semantic similarity via deep interaction and shows much better performance than the dual-encoder architecture~\cite{luan2020sparse}. The cross-encoder is more effective and robust, while it is inefficient over a large number of candidates in inference. Hence, we first train a cross-encoder (following the architecture shown in Figure~\ref{fig-cross-encoder}). Then, when sampling hard negatives from the top-ranked passages retrieved by a dense retriever, we select only the passages that are predicted as negatives by the cross-encoder with high confidence scores. The selected top-retrieved passages can be considered as denosied samples that are more reliable to be used as hard negatives. 

\textbf{Data Augmentation} The third strategy aims to alleviate the issue of limited training data. Since the cross-encoder is more powerful in measuring the similarity between questions and passages, we utilize it to annotate unlabeled questions for data augmentation. 
Specifically, we incorporate a new collection of unlabeled questions, while reuse the passage collection. 
Then, we use the learned cross-encoder to predict the passage labels for the new questions. 
To ensure the quality of the automatically labeled data, we only select the predicted positive and negative passages with high confidence scores estimated by the cross-encoder. 
 Finally, the automatically labeled data is used as augmented training data to learn the dual encoder. Another view of the data augmentation is knowledge distillation~\citep{Hinton2015DistillingTK}, where the cross-encoder is the teacher and the dual-encoder is the student.

\begin{table*}[h]
\centering
\small
\begin{tabular}{ccccccc}
\hline 
\textbf{datasets} & \textbf{\#q in train} & \textbf{\#q in dev} & \textbf{\#q in test} & \textbf{\#p} & \textbf{ave. q length} & \textbf{ave. p length} \\
\hline
MSMARCO & 502,939              & 6,980              & 6,837                 & 8,841,823  & 5.97            & 56.58          \\
\hline
NQ     & 58,812       & \multicolumn{1}{c}{-}    & 3,610  & 21,015,324 & 9.20            & 100.0  \\
\hline       
\end{tabular}
\centering
\caption{The statistics of datasets MSMARCO and Natural Questions. Here, ``p'' and ``q'' are the abbreviations of  questions and passages, respectively. The length is in tokens. }
\label{tbl-datastat}
\vspace{-3mm}
\end{table*}

\subsection{The Training Procedure}
As shown in Figure~\ref{fig-model}, we organize the above three training strategies into an effective training pipeline for the dual-encoder. It makes an analogy to a multi-stage rocket, where the performance of the dual-encoder is consecutively improved at three steps (STEP 1, 3 and 4). That is why we call our approach \emph{RocketQA}. Next, we will describe the details of the whole training procedure of RocketQA. 
\begin{itemize}[noitemsep,topsep=0pt,parsep=0pt,partopsep=0pt,labelindent=0cm,leftmargin=0cm]
	\item \textbf{REQUIRE:} Let $C$ denote a collection of passages. $Q_L$ is a set of questions that have corresponding labeled passages in $C$, and $Q_U$ is a set of questions that have no corresponding labeled passages. $D_L$ is a dataset consisting of $C$ and $Q_L$, and $D_U$ is a dataset consisting of $C$ and $Q_U$. 
	\item \textbf{STEP 1:} Train a dual-encoder $M_D^{(0)}$ by using cross-batch negatives on $D_L$. 
	\item \textbf{STEP 2:} Train a cross-encoder $M_C$ on $D_L$. The positives used for training the cross-encoder are from the original training set $D_L$, while the negatives are randomly sampled from the top-$k$ passages (excluding the labeled positive passages) retrieved by $M_D^{(0)}$ from $C$ for each question $q \in Q_L$. This design is to let the cross-encoder adjust to the distribution of the results retrieved by the dual-encoder, since the cross-encoder will be used in the following two steps for optimizing the dual-encoder. This design is important, and there is similar observation in Facebook Search~\citep{huang2020embedding}. 
	\item \textbf{STEP 3:} Train a dual-encoder $M_D^{(1)}$ by further introducing denoised hard negative sampling on $D_L$. Regarding to each question $q \in Q_L$, the hard negatives are sampled from the top passages retrieved by $M_D^{(0)}$ from $C$, and only the passages that are predicted as negatives by the cross-encoder $M_C$ with high confidence scores will be selected. 
	\item \textbf{STEP 4:} Construct pseudo training data $D_U$ by using $M_C$ to label the top-$k$ passages retrieved by $M_D^{(1)}$ from $C$ for each question $q \in Q_U$, and then train a dual-encoder $M_D^{(2)}$ on both the manually labeled training data $D_L$ and the automatically augmented training data $D_U$. 
\end{itemize}

Note that the cross-batch negative strategy is applied through all the steps for training the dual-encoder. The cross-encoder is used both STEP 3 and STEP 4 with different purposes to promote the performance of the dual encoder. The implementation details of denoising hard negatives and data augmentation can be found in Section ~\ref{experiments}.

\section{Experiments} \label{experiments}

\subsection{Experimental Setup}

\subsubsection{Datasets}
We conduct the experiments on two popular QA benchmarks: MSMARCO Passage Ranking \citep{nguyen2016ms} and Natural Questions (NQ) \citep{kwiatkowski2019natural}. 
The statistics of the datasets are listed in Table~\ref{tbl-datastat}. 

\textbf{MSMARCO Passage Ranking} MSMARCO is originally designed for multiple passage MRC, and its questions were sampled from Bing search logs. Based on the questions and passages in MSMARCO Question Answering, a dataset for passage ranking was created, namely MSMARCO Passage Ranking, consisting of about $8.8$ million passages. The goal is to find positive passages that answer the questions. 

\textbf{Natural Question (NQ)} \citet{kwiatkowski2019natural} introduces a large dataset for open-domain QA. The original dataset contains more than $300,000$ questions collected from Google search logs. In \citet{karpukhin2020dense},  around $62,000$ factoid questions are selected, and all the  Wikipedia articles  are processed as the collection of passages. There are more than $21$ million passages in the corpus. In our experiments, we reuse the version of NQ created by \citet{karpukhin2020dense}. Note that the dataset used in DPR contains empty negatives, and we discarded the empty ones.

\subsubsection{Evaluation Metrics}
Following previous work, we use MRR and Recall at top $k$ ranks to evaluate the performance of passage retrieval, and exact match (EM) to measure the performance of answer extraction. 

\textbf{MRR} The Reciprocal Rank (RR) calculates the reciprocal of the rank at which the first relevant passage was retrieved. When averaged across questions, it is called Mean Reciprocal Rank (MRR).

\textbf{Recall at top $k$ ranks} The top-$k$ recall of a retriever is defined as the proportion of questions to which the top $k$ retrieved passages contain answers. 

\textbf{Exact match} This metric measures the percentage of questions whose predicted answers that match any one of the reference answers exactly, after string normalization. 

\begin{table*}
\centering
\small
\begin{tabular}{cccccccc}
                     \hline           
            \multirow{2}*{\textbf{Methods}}         & 
            \multirow{2}*{\textbf{PLMs}} & 
            \multicolumn{3}{c}{\textbf{MSMARCO Dev}} & \multicolumn{3}{c}{\textbf{Natural Questions Test}} \\
                    &  & MRR@10    & R@50    & R@1000    & R@5         & R@20         & R@100         \\
                     \hline
BM25 (anserini)~\citep{yang2017anserini} & -     & 18.7      & 59.2    & 85.7      &    -         &   59.1           &       73.7        \\ \hline
doc2query~\citep{nogueira2019document} & -            & 21.5      & 64.4    & 89.1      &      -       &     -         &     -         \\
DeepCT~\citep{dai2019deeper} & -               & 24.3      & 69.0    & 91.0      &      -       &     -         &     -         \\
docTTTTTquery~\citep{nogueira2019doc2query} & -        & 27.7      & 75.6    & 94.7      &     -        &      -        &     -         \\
GAR~\citep{mao2020generation} & -        & -      & -       & -      & -           & 74.4         & 85.3  \\
\hline
DPR (single)~\citep{karpukhin2020dense} & BERT$_\text{base}$         & -         & -       & -         & -           & 78.4         & 85.4          \\
ANCE (single)~\citep{xiong2020approximate} & RoBERTa$_\text{base}$        & 33.0      & -       & 95.9      & -           & 81.9         & 87.5          \\
ME-BERT~\citep{luan2020sparse} & BERT$_\text{large}$              & 33.8      & -       & -         & -           & -            & -             \\
                     \hline
RocketQA & ERNIE$_\text{base}$                                                                  & \textbf{37.0}     & \textbf{85.5}   & \textbf{97.9}     & \textbf{74.0}       & \textbf{82.7}       & \textbf{88.5}        \\
                     \hline           
\end{tabular}
\centering
\caption{The performance comparison on passage retrieval. Note that we directly copy the reported numbers from the original papers and leave the blanks if they were not reported. }
\label{tbl-mainres}
\vspace{-4mm}
\end{table*}

\subsubsection{Implementation Details}
We conduct all experiments with the deep learning framework PaddlePaddle~\citep{Ma2019PaddlePaddleAO} on up to eight NVIDIA Tesla V100 GPUs (with 32G RAM). 

\textbf{Pre-trained LMs} The dual-encoder is initialized with the parameters of ERNIE 2.0 base ~\citep{sun2019ernie}, and the cross-encoder is initialized with ERNIE 2.0 large. ERNIE 2.0 has the same networks as BERT, and it introduces continual pre-training framework on multiple pre-trained tasks. We notice previous work use different pre-trained LMs, and we examine the effects of pre-trained LMs in Section ~\ref{plm-effect} in Appendix. Our approach is effective when using different pre-trained LMs. 

\textbf{Cross-batch negatives ~\footnote{When using multi-GPUs, the cross-batch negatives is as efficient as the in-batch negatives. Because the cross-batch re-uses the computed embeddings of paragraphs and the communication cost of embeddings across GPUs can be negligible.}} 
The cross-batch negative sampling is implemented with differentiable all-gather operation provided in FleetX~\citep{dong2020}, that is a highly scalable distributed training engine of PaddlePaddle. The all-gather operator makes representation of passages across all GPUs visible on each GPU and thus the cross-batch negative sampling approach can be applied globally. 

\textbf{Denoised hard negatives and data augmentation} We use the cross-encoder for both denoising hard negatives and data augmentation. Specifically, we select the top retrieved passages with scores less than $0.1$ as negatives and those with scores higher than $0.9$ as positives. We manually evaluated the selected data, and the accuracy was higher than $90\%$. 


\textbf{The number of positives and negatives}  When training the cross-encoders, the ratios of the number of positives to the number of negatives are 1:4 and 1:1 on MSMARCO and NQ, respectively. The negatives used for training cross-encoders are randomly sampled from top-$1000$ and top-$100$ passages retrieved by the dual-encoder $M_D^{(0)}$ on MSMARCO and NQ, respectively. 
When training the dual-encoders in the last two steps ($M_D^{(1)}$ and $M_D^{(2)}$), we set the ratios of the number of positives to the number of hard negatives as 1:4 and 1:1 on MSMARCO and NQ, respectively.

\textbf{Batch sizes} The dual-encoders are trained with the batch sizes of $512 \times 8$ and $512 \times 2$ on MSMARCO and NQ, respectively. The batch size used on MSMARCO is larger, since the size of MSMARCO is larger than NQ.  The cross-encoders are trained with the batch sizes of $64 \times 4$ and $64$ on MSMARCO and NQ, respectively. We use the automatic mixed precision and gradient checkpoint~\footnote{The gradient checkpoint~\citep{chen2016training} enables the trading off computation against memory resulting in sublinear memory cost, so  bigger/deeper nets can be trained with limited resources. } functionality in FleetX, so as we can train the models using large batch sizes with limited resources. 

\textbf{Training epochs} The dual-encoders are trained on MSMARCO for $40$, $10$ and $10$ epochs in three steps of RocketQA, respectively. 
The dual-encoders are trained on NQ for $30$ epochs in all steps of RocketQA. 
The cross-encoders are trained for $2$ epochs on both MSMARCO and NQ. 

\textbf{Optimizers} We use ADAM optimizer. 

\textbf{Warmup and learning rate} The learning rate of the dual-encoder is set to 3e-5 and the rate of linear scheduling warm-up is set to $0.1$, while the learning rate of the cross-encoder is set to 1e-5. 

\textbf{Maximal length} We set the maximal length of questions and passages as 32 and 128, respectively.

\textbf{Unlabeled questions}
We collect $1.7$ million unlabeled questions from Yahoo! Answers\footnote{\scriptsize \url{http://answers.yahoo.com/}}, ORCAS \citep{craswell2020orcas} and MRQA~\citep{Fisch2019MRQA2S}. We use the questions from Yahoo! Answers, ORCAS and NQ as new questions in the experiments of MSMARCO. We only use the questions from MRQA as the new questions in the experiments of NQ. Since both NQ and MRQA mainly contain factoid-questions, while other datasets contain both factoid and non-factoid questions.

\subsection{Experimental Results}

In our experiments, we first examine the effectiveness of our retriever on MSMARCO and NQ datasets. Then, we conduct extensive experiments to examine the effects of the three proposed training strategies. We also show the performance of end-to-end QA based on our retriever on NQ dataset. 

\subsubsection{Dense Passage Retrieval}

We first compare RocketQA with the previous state-of-the-art approaches on passage retrieval. 
We consider both sparse and dense passage retriever baselines.
The sparse retrievers include the traditional retriever BM25~\citep{yang2017anserini}, and four traditional retrievers enhanced by neural networks, including doc2query~\citep{nogueira2019document}, DeepCT~\citep{dai2019deeper}, docTTTTTquery~\citep{nogueira2019doc2query} and GAR~\citep{mao2020generation}. Both doc2query and docTTTTTquery employ neural question generation to expand documents. In contrast, GAR employs neural generation models to expand questions.  Different from them, DeepCT utilizes BERT to learn the term weight. 
The dense passage retrievers include DPR~\citep{karpukhin2020dense}, ME-BERT~\citep{luan2020sparse} and ANCE~\citep{xiong2020approximate}. Both DRP and ME-BERT use in-batch random sampling and hard negative sampling from the results retrieved by BM25, while ANCE enhances the hard negative sampling by using the dense retriever. 

Table~\ref{tbl-mainres} shows the main experimental results. We can see that RocketQA significantly outperforms all the baselines on both MSMARCO and NQ datasets.
Another observation is that the dense retrievers are overall better than the sparse retrievers. Such a finding has also been reported in previous studies~\citep{karpukhin2020dense,luan2020sparse,xiong2020approximate}, which indicates the effectiveness of the dense retrieval approach.

\begin{table}[]
\centering
\small
\begin{tabular}{lcc}
\hline
\textbf{Strategy}                 & \textbf{MRR@10} \\ \hline
In-batch negatives         & 32.39           \\
Cross-batch negatives (i.e. STEP 1)     & 33.32           \\ \hline
Hard negatives w/o denoising & 26.03            \\
Hard negatives w/ denoising (i.e. STEP 3)  & 36.38            \\ 
\hline
Data augmentation (i.e. STEP 4)        & \textbf{37.02}  \\ \hline
\end{tabular}
\centering
\caption{The experiments to examine the effectiveness of the three proposed training strategies in RocketQA on MSMARCO Passage Ranking. }
\label{tbl-ablation}
\end{table}

\begin{figure}[]
	\centering 
	\includegraphics[width=0.42\textwidth]{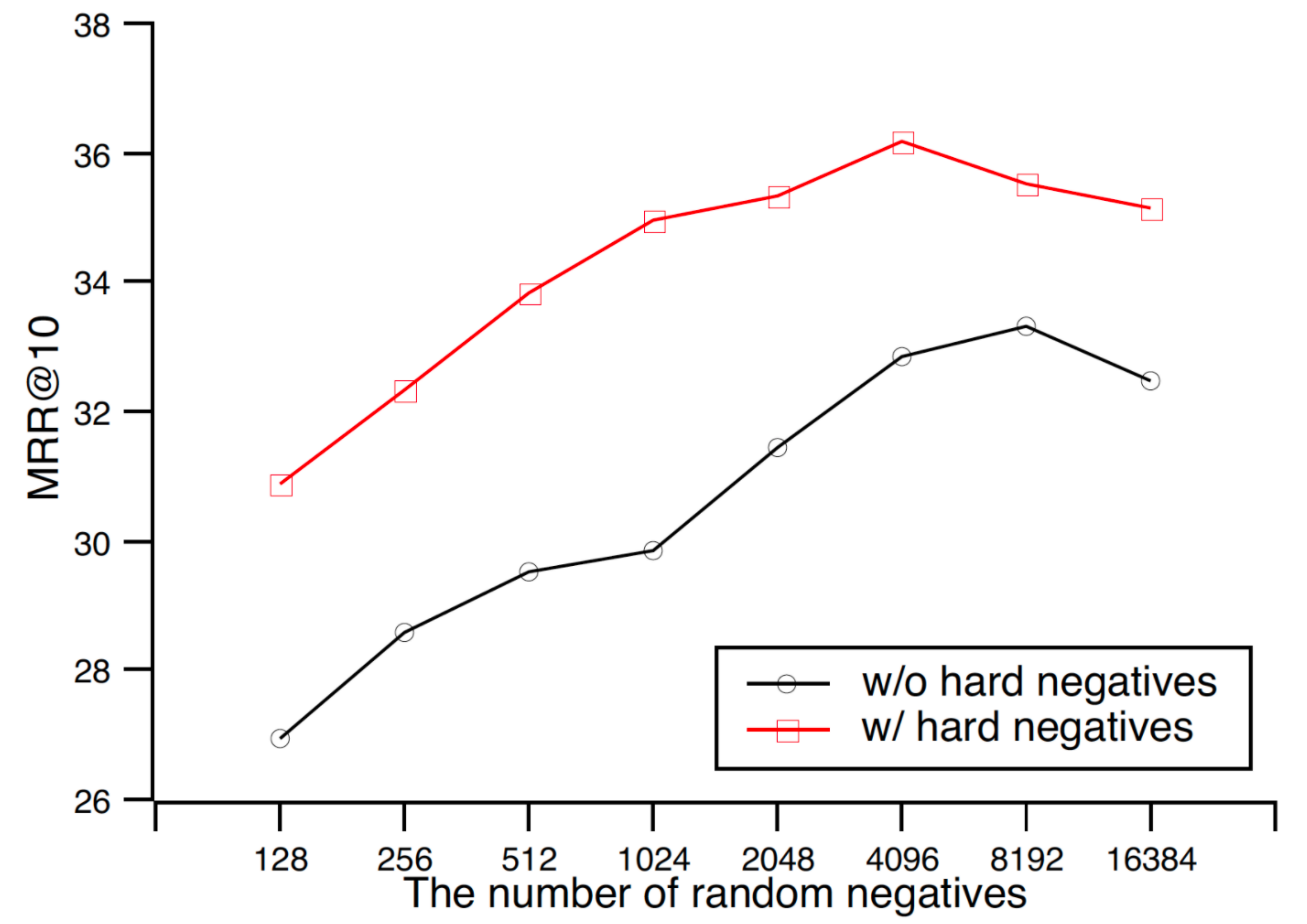}
	\vspace{-3mm}
	\caption{The effect of the number of random negatives paired for a question on MSMARCO dataset. The models without and with hard negatives are trained with 20K and 5K steps, respectively. } 
	\label{fig-batch-size-effect} 
\vspace{-3mm}
\end{figure}

\subsubsection{The Effectiveness of The Three Training Strategies in RocketQA}
In this part, we conduct the extensive experiments on MSMARCO dataset to examine the effectiveness of the three strategies in RocketQA. Results on NQ dataset has shown the similar findings (see in Section ~\ref{effect-nq} in Appendix).

First, we compare cross-batch negatives with in-batch negatives by using the same experimental setting (i.e. the number of epochs is $40$ and the batch size is $512$ on each single GPU). From the first two rows in Table~\ref{tbl-ablation}, we can see that the performance of the dense retriever can be improved with more negatives by cross-batch negatives.  It is expected that when increasing the number of random negatives, it will reduce the discrepancy between training and inference. Furthermore, we investigate the effect of the number of random negatives. Specifically, we examine the performance of dual-encoders trained by using different numbers of random negatives with a fixed number of steps. From Figure~\ref{fig-batch-size-effect}, we can see that the model performance increases, when the number of random negatives becomes larger. After a certain point, the model performance starts to drop, since a large batch size may bring difficulty for optimization on training data with limited size. 
We say that there should be a balance between the batch size and the number of negatives. When increasing the batch size, we will have more negatives for each question. However, when the size of training data is limited, a large batch size will bring difficulty for optimization.

\begin{figure}[!tb]
	\centering 
	\includegraphics[width=0.42\textwidth]{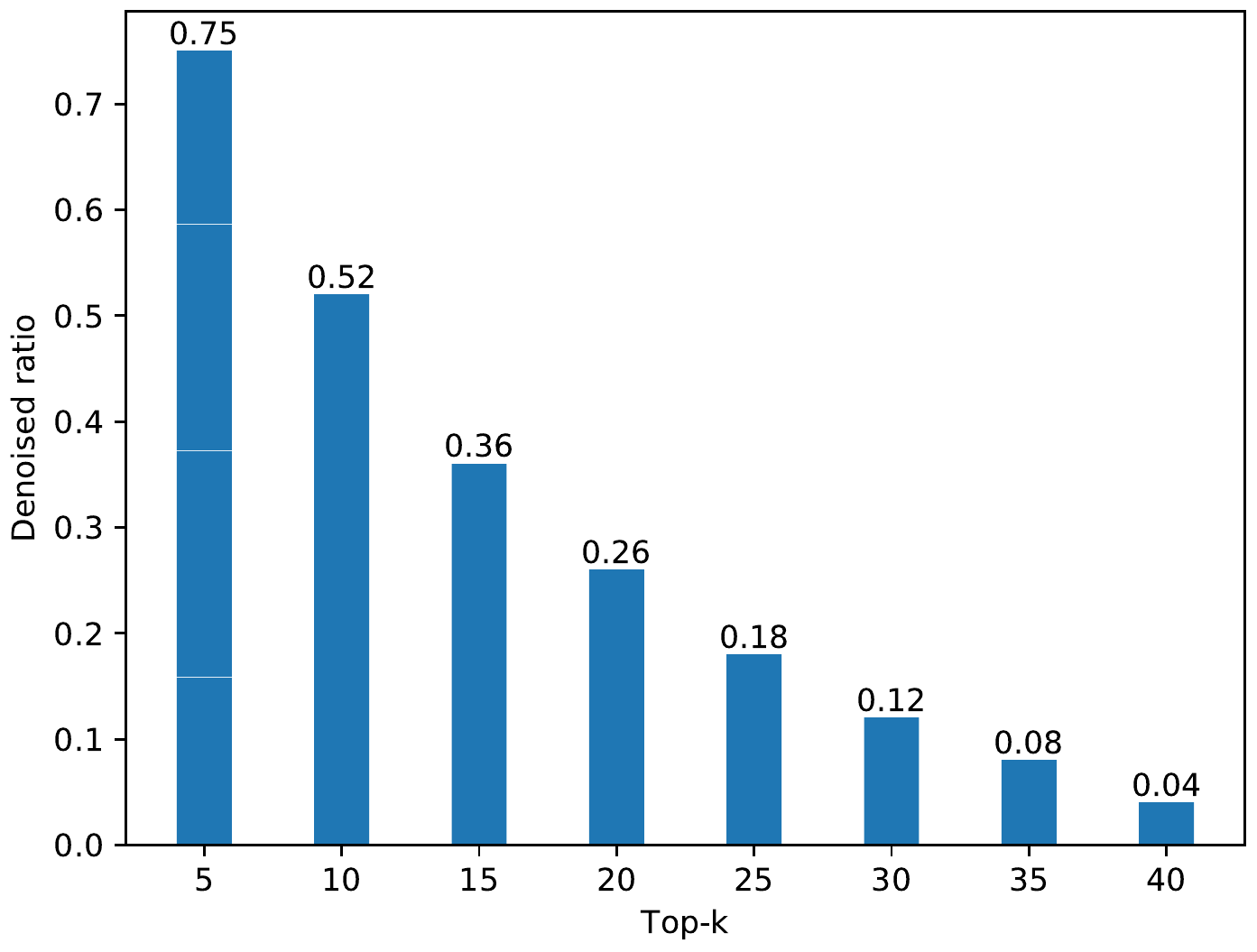}
	\vspace{-3mm}
	\caption{The ratios of denoised passages at different ranks on MSMARCO. }	\label{fig-denoised-ratio} 
\vspace{-3mm}
\end{figure}

\begin{table*}[!t]
\tiny
\begin{tabular}{p{2.1cm}|p{3.5cm}|p{4.4cm}|p{4.4cm}}

\toprule
\textbf{Question} & \textbf{Label positives} & \textbf{Hard negatives w/o denoising (false negatives)} & \textbf{Hard negatives w/ denoising} \\
\midrule
\multirow{1}{2.1cm}[-1em]{How many kilohertz in a megahertz} 
&  {One \textbf{megahertz (abbreviated: MHz) is equal to 1,000 kilohertz, or 1,000,000 hertz}. It can also be described as one million cycles per second. \ldots}
& {(\textit{\underline{Rank 2nd}}) Kilo means times 1000, mega means times 1,000,000. So \textbf{0.005 megahertz} = 5000 Hz = \textbf{5 kiloHz}. Hertz (not Herz) is abbreviated to Hz.  \ldots} 
& {(\textit{\underline{Rank 14th}}) \ldots \textbf{megahertz (MHz) and gigahertz (GHz)} are used to measure CPU speed. For example, a 1.6 GHz computer processes data internally \ldots}
\\
\midrule
\multirow{1}{2.1cm}[-1em]{Name of test for achilles tendon rupture} 
& {In a patient with a \textbf{ruptured Achilles tendon}, the foot will not move. That is called a positive \textbf{Thompson test}. The Thompson test is important because\ldots}
& {(\textit{\underline{Rank 1st}}) \ldots The physical examination should include two or more of the following tests to establish the diagnosis of acute \textbf{Achilles tendon rupture}: Clinical \textbf{Thompson test} \ldots} 
& {(\textit{\underline{Rank 9th}}) \ldots Methods: Ultrasound was used to measure \textbf{Achilles tendon}. length and muscle-tendon architectural parameters in children. of ages 5 to 12 years. \ldots}
\\
\bottomrule
\end{tabular}
\caption{The hard negatives before and after denoising on MSMARCO. The bolded words are the keywords relevant to questions. }
\label{tbl-denoising}
\end{table*}

Second, we examine the effect of denoised hard negatives from the top-$k$ passages retrieved by the dense retriever. 
As shown in the third row in Table~\ref{tbl-ablation},  the performance of the retriever significantly decreases by introducing hard negatives without denoising. 
We speculate that it is caused by the fact that there are a large number of unlabeled positives. 
Specifically, we manually examine the top-retrieved passages of $100$ questions, that were not labeled as true positives. We find that about $70\%$ of them are actually positives or highly relevant. Hence, it is likely to bring noise if we simply sample hard negatives from the top-retrieved passages by the dense retriever, which is a widely adopted strategy to sample hard negatives in previous studies~\citep{gillick2019learning,wu2019zero,xiong2020approximate}. 
As a comparison, we propose denoised hard negatives by a powerful cross-encoder. From the fourth row in Table~\ref{tbl-ablation}, we can see that denoised negatives improve the performance of the dense retriever. 
To obtain more insights about denoised hard negatives, Table \ref{tbl-denoising} gives the sampled hard negatives for two questions before and after denoising. 
Figure ~\ref{fig-denoised-ratio} further illustrates the ratio of filtered passages at different ranks. We can see that there are more passages filtered (i.e. denoised) at lower ranks, since it is likely to have more false negatives at lower ranks.

\begin{figure}[!tb]
	\centering 
	\includegraphics[width=0.42\textwidth]{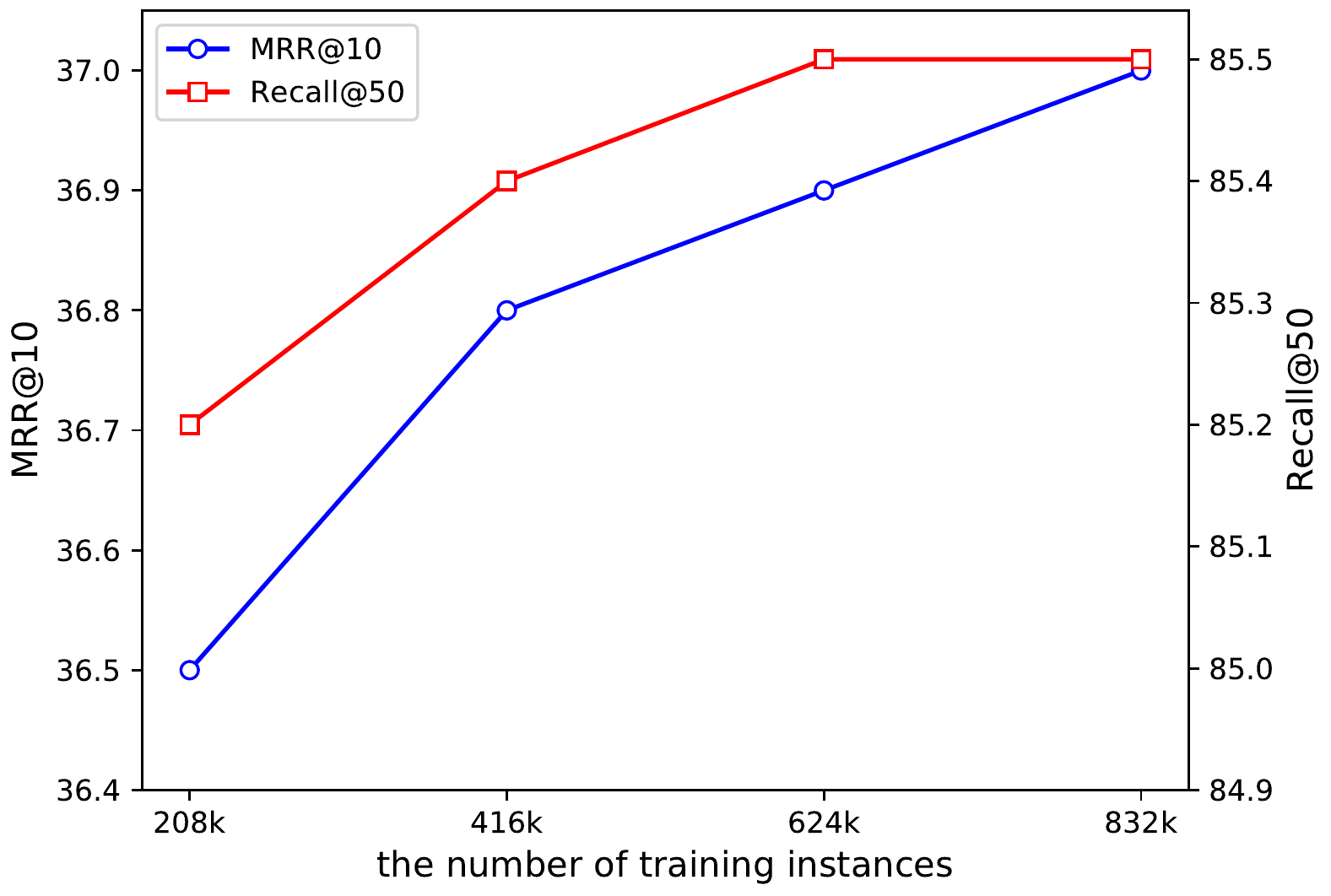}
	\vspace{-3mm}
	\caption{The effect of the size of augmented data on MSMARCO dataset. } 
	\label{fig-size-dataaug} 
\vspace{-2mm}
\end{figure}

Finally, when integrated with the data augmentation strategy (see the fifth row in Table~\ref{tbl-ablation}), the performance has been further improved. A major merit of data augmentation is that it does not explicitly rely on manually-labeled data. Instead, it utilizes the cross-encoder (having more powerful capability than the dual-encoder) to generate pseudo training data for improving the dual-encoder. We further examine the effect of the size of the augmented data. As shown in Figure ~\ref{fig-size-dataaug}, we can see when the size of the augmented data is increasing, the performance increases. 

\subsubsection{Passage Reading with RocketQA}
Previous experiments have shown the effectiveness of RocketQA on passage retrieval. Next, we  verify whether the retrieval results of RocketQA can improve the performance of passage reading for extracting correct answers. We implement an end-to-end QA system in which we have an extractive reader stacked on our RocketQA retriever. For a fair comparison, we first re-use the released model~\footnote{\url{https://github.com/facebookresearch/DPR}} of the extractive reader in DPR~\citep{karpukhin2020dense}, and take 100 retrieved passages during inference (the same setting used in DPR). Besides, we use the same setting to train a new extractive reader based on the retrieval results of RocketQA (except that we choose top 50 passages for training instead of 100). The motivation is that the reader should be adapted to the retrieval distribution of RocketQA. 

Table~\ref{tbl-nqmrc} summarizes the the end-to-end QA performance of our approach and a number of competitive methods. 
From Table~\ref{tbl-nqmrc}, we can see that our retriever leads to better QA performance. Compared with prior solutions, our novelty mainly lies in the passage retrieval component, i.e., the RocketQA approach. The results have shown that our approach can provide better passage retrieval results, which finally improve the final QA performance.  

\begin{table}
\centering
\small
\begin{tabular}{ccc}
\hline
\textbf{Model}                                & \textbf{EM}   \\ \hline
BM25+BERT~\cite{lee2019latent}       & 26.5 \\ 
HardEM~\cite{min2019discrete}         & 28.1 \\ 
GraphRetriever~\cite{min2019knowledge} & 34.5 \\ 
PathRetriever~\cite{asai2019learning}  & 32.6 \\ 
ORQA~\cite{lee2019latent}            & 33.3 \\ 
REALM~\cite{guu2020realm}            & 40.4 \\ 
DPR~\cite{karpukhin2020dense}        & 41.5 \\ 
GAR~\cite{mao2020generation} & 41.6 \\ \hline
RocketQA + DPR reader                                & 42.0  \\ 
RocketQA + re-trained DPR reader                                & \textbf{42.8}  \\ \hline
\end{tabular}
\centering
\caption{The experimental results of passage reading on NQ dataset. In this paper, we focus on extractive reader, while the recent generative readers~\citep{lewis2020retrieval,izacard2020leveraging} can also be applied here and may lead to better results. }
\label{tbl-nqmrc}
\vspace{-4mm}
\end{table}

\section{Conclusions}
In this paper, we have presented an optimized training approach to improving dense passage retrieval. We have made three major technical contributions in RocketQA, namely cross-batch negatives, denoised hard negatives and data augmentation. 
Extensive experiments have shown the effectiveness of the proposed approach by incorporating the three  optimization strategies. We also demonstrate that  the performance of end-to-end QA can be improved based on our RocketQA retriever. 

\section{Ethical Considerations}

The technique of dense passage retrieval is effective for question answering, where the majority of questions are informational queries. 
Different from the traditional search, there is usually term mismatch between questions and answers. The term mismatch brings barriers for the machine to accurately find the information for people. 
Hence, we need dense passage retrieval for semantic matching in the scenario of question answering. 
Dense passage retrieval has the potential to empower people to find the accurate information more quickly and achieve more in their daily life and work. 
Our technique contributes toward the goal of asking machines to find the answers to natural language questions from a large collection of documents. However, the goal is still far from being achieved, and more efforts from the community is needed for us to get there.

\section{Acknowledgments}

This work is supported by the National Key Research and Development Project of China (No. 2018AAA0101900). We would also like to thank the anonymous reviewers for their insightful suggestions.


\bibliography{custom}
\bibliographystyle{acl_natbib}

\clearpage
\appendix

\section{Appendix}
\label{sec:appendix}

\subsection{The Effects of Pre-trained LMs} \label{plm-effect}
We notice that previous work use different pre-trained LMs. As shown in Table~\ref{tbl-plm-effect}, DPR~\citep{karpukhin2020dense} uses BERT\textsubscript{base}. ANCE~\citep{xiong2020approximate} uses RoBERTa\textsubscript{base}, and ME-BERT~\citep{luan2020sparse} uses BERT\textsubscript{large}. We mainly use ERNIE\textsubscript{base} in our experiments. In this section, we try to examine the effects of pre-trained LMs for RocketQA. 
Specifically, we use BERT\textsubscript{base} to replace ERNIE\textsubscript{base}, and apply it to the first step of RocketQA. 
From Table~\ref{tbl-plm-effect} (see the forth row and the fifth row), we can observe that the performance slightly decreases when using BERT\textsubscript{base}. 
In other words, comparing to BERT\textsubscript{base}, ERNIE\textsubscript{base} brings gains about $0.6$ in terms of MRR@10 on MSMARCO, and $1.6$ in terms of R@100 on NQ, respectively. 
However, RocketQA trained only with cross-batch negatives is already comparable to previous work, including DPR, ANCE and ME-BERT (although they employ better pre-trained LMs). 
We conclude that our approach is still effective when using different pre-trained LMs.

\begin{table}[htb]
\centering
\small
\begin{tabular}{cccc}
\hline
\multicolumn{1}{c}{\multirow{2}{*}{\textbf{Methods}}} & \multirow{2}{*}{\textbf{PLMs}} & \textbf{MSMARCO} & \textbf{NQ} \\
\multicolumn{1}{c}{}                         &                                  & MRR@10  & R@100             \\ \hline
DPR (single)  &    BERT\textsubscript{base}   &  -       &     85.4              \\
ANCE (single)   &   RoBERTa\textsubscript{base}   &         33.0 &       87.5            \\
ME-BERT    &     BERT\textsubscript{large}  &  33.8       &       -            \\ \hline
RocketQA\textsubscript{STEP1}      &    BERT\textsubscript{base} & 32.7        &           86.0        \\
RocketQA\textsubscript{STEP1}  &  ERNIE\textsubscript{base} &  33.3       & 87.6 \\
RocketQA &  ERNIE\textsubscript{base} &    37.0     &        88.5           \\  \hline
\end{tabular}
\centering
\caption{The effects of pre-trained LMs. Note that we directly copy the reported numbers from the original papers and leave the blanks if they were not reported. }
\label{tbl-plm-effect}
\end{table}

\subsection{The Effectiveness of The Three Training Strategies on NQ} \label{effect-nq}

In this section, we examine the effectiveness of the three proposed training strategies on NQ dataset. From Table ~\ref{tbl-ablation-nq}, we can observe that all the three strategies are effective. The findings are similar to the results on MSMARCO.  

\begin{table}[htb]
\centering
\small
\begin{tabular}{lcc}
\hline
\textbf{Strategy}                 & \textbf{R@5} \\ \hline
In-batch negatives         & 68.5           \\
Cross-batch negatives (i.e. STEP 1)     & 68.9           \\ \hline
Hard negatives w/o denoising & 68.0            \\
Hard negatives w/ denoising (i.e. STEP 3)  & 73.2            \\ 
\hline
Data augmentation (i.e. STEP 4)        & \textbf{74.0}  \\ \hline
\end{tabular}
\centering
\caption{The experiments to examine the effectiveness of the three proposed training strategies in RocketQA on NQ. }
\label{tbl-ablation-nq}
\end{table}

\end{document}